\documentclass[12pt]{article}
\usepackage{amsmath}
\usepackage{times}
\usepackage{graphicx}
\usepackage{color}
\usepackage{multirow}
\usepackage[authoryear]{natbib}
\usepackage{rotating}
\usepackage{bbm}
\usepackage{latexsym}
\usepackage{helvet}
\usepackage{courier}
\usepackage{graphicx}
\usepackage{booktabs}
\usepackage{amssymb}
\usepackage{subfigure}
\usepackage{algorithm}
\usepackage{algorithmic}

\def\A{{\bf A}}

\def\f{{\bf f}}

\def\Y{{\bf Y}}

\def\y{{\bf y}}
\def\z{{\bf z}}
\def\Z{{\bf Z}}
\def\M{{\bf M}}

\def\U{{\bf U}}

\def\V{{\bf V}}

\def\W{{\bf W}}

\def\0{{\bf 0}}
\def\1{{\bf 1}}

\def\AM{{\mathcal A}}
\def\BM{{\mathcal B}}

\def\YM{{\mathcal Y}}

\def\LM{{\mathcal L}}

\def\RB{{\mathbb R}}

\def\tr{\mathrm{tr}}

\newtheorem{remark}{Remark}
\newtheorem{theorem}{Theorem}
\newtheorem{lemma}{Lemma}
\newtheorem{definition}{Definition}

\newtheorem{cor}{Corollary}

\def\proof{\textbf{Proof: }}
\def\endproof{$\blacksquare$}

\DeclareGraphicsExtensions{.png}

\newcommand{\captionfonts}{\normalsize}

\makeatletter
\long\def\@makecaption#1#2{%
  \vskip\abovecaptionskip
  \sbox\@tempboxa{{\captionfonts #1: #2}}%
  \ifdim \wd\@tempboxa >\hsize
    {\captionfonts #1: #2\par}
  \else
    \hbox to\hsize{\hfil\box\@tempboxa\hfil}%
  \fi
  \vskip\belowcaptionskip}
\makeatother

\begin{document}
\hspace{13.9cm}1

\ \vspace{20mm}\\

{\LARGE Large Margin Low Rank Tensor Analysis}

\ \\
{\bf \large Guoqiang~Zhong and
        Mohamed~Cheriet}\\
Synchromedia Laboratory for Multimedia Communication in Telepresence, \'{E}cole de Technologie Sup\'{e}rieure,
    Montr\'{e}al, Qu\'{e}bec H3C 1K3, Canada. \protect  \\
    E-mail: guoqiang.zhong@synchromedia.ca, mohamed.cheriet@etsmtl.ca.\\
%


\thispagestyle{empty}
\markboth{}{NC instructions}
\ \vspace{-0mm}\\
%
\begin{center} {\bf Abstract} \end{center}
Other than vector representations, the direct objects of human cognition are generally high-order tensors, such as 2D images and 3D textures. From this fact, two interesting questions naturally arise: How does the human brain represent these tensor perceptions in a ``manifold" way, and how can they be recognized on the ``manifold"? In this paper, we present a supervised model to learn the intrinsic structure of the tensors embedded in a high dimensional Euclidean space. With the fixed point continuation procedures, our model automatically and jointly discovers the optimal dimensionality and the representations of the low dimensional embeddings. This makes it an effective simulation of the cognitive process of human brain. Furthermore, the generalization of our model based on similarity between the learned low dimensional embeddings can be viewed as counterpart of recognition of human brain. Experiments on applications for object recognition and face recognition demonstrate the superiority of our proposed model over state-of-the-art approaches.


\section{Introduction}

In one paper by~\cite{Seung22122000}, the authors state that the human brain represents real world perceptual stimuli in a manifold way -- encoding high dimensional signals in an intrinsically low dimensional structure. At the same time of their work and later on, numerous manifold learning algorithms, such as isometric feature mapping (Isomap)~\citep{Tenenbaum22122000} and locally linear embedding (LLE)~\citep{Roweis22122000}, were proposed for discovering the manifold structure of data embedded in a high dimensional space. Most of these manifold learning methods can be applied to vector representations of signals, and yield acceptable performance for visualization and recognition. However, in contrast, humans can perceive not only vector representations of signals (one-order tensors), but also high order representations (high-order tensors), such as 2D images and 3D textures. More importantly, humans can in general perform high accuracy recognition based on learned patterns, i.e. recognizing objects and faces. From this fact, two questions naturally arise: How does the human brain learn the intrinsic manifold structure of the tensor representations, and how does it recognize new patterns based on the learned manifold structure?

To solve these two questions, some researchers try to extend traditional vector representation-based dimensionality reduction approaches to the applications related to high order tensors. Specifically, some representative tensor dimensionality reduction approaches include~\citep{YangZFY04, YeJL04} and~\citep{WangYHT07}. These approaches can learn the low dimensional representations of tensors in either an unsupervised or a supervised way. In particular, the approach presented in~\citep{WangYHT07} is theoretically guaranteed to converge to a local optimal solution of the learning problem. However, one common issue of these approaches exists: the dimensionality of the low dimensional tensor space must be manually specified before these approaches are applied. Therefore, these approaches may not necessarily lead to the genuine manifold structure of the tensors.

To exploit the questions above and overcome the shortage of previous approaches, in this paper, we propose a novel tensor dimensionality reduction method, called large margin low rank tensor analysis (LMLRTA). LMLRTA is aimed at learning the low dimensional representations of tensors using techniques of multi-linear algebra~\citep{Algebra} and graph theories~\citep{Bondy}. Compared to traditional vector representation-based dimensionality reduction approaches, LMLRTA can take any order of tensors as input, including 1D vectors (one-order tensor), 2D matrices (two-order tensor), and more. This guarantees the feasibility of that one can use LMLRTA to simulate the way how human brain represents perceived signals, such as speech, images and textures. Furthermore, unlike previous tensor dimensionality reduction approaches~\citep{YangZFY04, YeJL04, WangYHT07}, which can only learn the low dimensional embeddings with a priori specified dimensionality, LMLRTA can automatically learn the optimal dimensionality of the tensor subspace. This guarantees LMLRTA to be an intelligent method to simulate the way of human perception. Besides, for the recognition of new coming patterns, we employ similarity between the learned low dimensional representations as measurement, which corresponds to the way how the human brain recognize new objects~\citep{Rosch}.

The rest of this paper is organized as follows. In Section 2, we provide an brief overview of previous work on dimensionality reduction. In Section 3, we present our proposed model, LMLRTA, in detail, including its formulation and optimization. Particularly, we theoretically prove that LMLRTA can converge to a local optimal solution of the optimization problem. Section 4 shows the experimental results on real world applications, including object recognition and face recognition, which are related to problems with respect to 2D tensors and 3D tensors, respectively. We conclude this paper in Section 5 with remarks and future work.

\section{Previous work}

In order to find the effective low dimensional representations of data, many dimensionality reduction approaches have been proposed in the areas of pattern recognition and machine learning. The most representative approaches are principal component analysis (PCA) and linear discriminant analysis (LDA) for the unsupervised and supervised learning paradigms, respectively. They are widely used in many applications due to their simplicity and efficiency. However, it is well known that both of them are optimal only if the relation between the latent and the observed space can be described with a linear function. To address this issue, nonlinear extensions based on kernel method have been proposed to provide nonlinear formulations, i.e. kernel principal component analysis (KPCA)~\citep{ScholkopfSM98} and generalized discriminant analysis (GDA)~\citep{Baudat2000}. 

Since about a decade ago, many manifold learning approaches have been proposed. These manifold learning approaches, including isometric feature mapping (Isomap)~\citep{Tenenbaum22122000} and locally linear embedding (LLE)~\citep{Roweis22122000}, can faithfully preserve global or local geometrical properties of the nonlinear structure of data. However, these methods only work on a given set of data points, and cannot be easily extended to out-of-sample data~\citep{BengioPVDRO03}. To alleviate this problem, locality preserving projections (LPP)~\citep{HeN03} and local fisher discriminant analysis (LFDA)~\citep{Sugiyama07} were proposed to approximate the manifold structure in a linear subspace by preserving local similarity between data points. In particular, Yan et al. proposed a general framework known as graph embedding for dimensionality reduction~\citep{YanXZZYL07}. Most of the spectral learning-based approaches, either linear or nonlinear, either supervised or unsupervised, are contained in this framework. Furthermore, based on this framework, the authors proposed the marginal Fisher analysis (MFA) algorithm for supervised linear dimensionality reduction. In the research of probabilistic learning models, \cite{Lawrence05} proposed the Gaussian process latent variable models (GPLVM), which extends PCA to a probabilistic nonlinear formulation. Combining a Gaussian Markov random field prior with GPLVM, \cite{ZhongLYHL10} proposed the Gaussian process latent random field model, which can be considered as a supervised variant of GPLVM. In the area of neural network research, ~\cite{Hinton2006} proposed a deep neural network model called autoencoder for dimensionality reduction. To exploit the effect of deep architecture for dimensionality reduction, some other deep neural network models were also introduced, such as deep belief nets (DBN)~\citep{HintonOT06}, stacked autoencoder (SAE)~\citep{BengioLPL06} and stacked denoise autoencoder (SDAE)~\citep{VincentLLBM10}. These studies show that deep neural networks can generally learn high level representations of data, which can benefit subsequent recognition tasks.

All of the above approaches assume that the input data are in the form of vectors. In many real world applications, however, the objects are essentially represented as high-order tensors, such as 2D images or 3D textures. One have to unfold these tensors into one-dimensional vectors first before the dimensionality reduction approaches can be applied. In this case, some useful information in the original data may not be sufficiently preserved. Moreover, high-dimensional vectorized representations suffer from the curse of dimensionality, as well as high computational cost. To alleviate these problems, 2DPCA~\citep{YangZFY04} and 2DLDA~\citep{YeJL04} were proposed to extend the original PCA and LDA algorithms to work directly on 2D matrices rather than 1D vectors. In recent years, many other approaches~\citep{YanXZZYL07, TaoTPAMI, Fu4429309, LiuLWY12, LiuLC10a} were also proposed to deal with high-order tensor problems. In particular, ~\cite{WangYHT07} proposed a tensor dimensionality reduction method based on the graph embedding framework, which is proved that it is the first method to give a convergent solution. However, as described before, all these previous tensor dimensionality reduction approaches have a common shortage: the dimensionality of the low dimensional representations must be specified manually before the approaches can be applied.

To address the above issues existing in both vector representation-based and tensor representation-based dimensionality reduction approaches, in this paper, we propose our novel method for tensor dimensionality reduction, called large margin low rank tensor analysis (LMLRTA). LMLRTA is able to take any order of tensors as input, and automatically learn the dimensionality of the low dimensional representations. More importantly, these merits make it an effective model to simulate the way how human brain represents and recognizes perceived signals.

\section{Large margin low rank tensor analysis (LMLRTA)}

In this section, we first introduce the used notation and some basic terminologies on tensor operations~\citep{KoldaB09, DYYeung}. And then, we detail our model, LMLRTA, including its formulation and optimization. Theoretical analyses to LMLRTA, such as its convergence, are also presented.

\subsection{Notation and terminologies}

We denote vector by using bold lowercase letter, such as $\bf{v}$, matrix by using bold uppercase letter, such as $\M$, and tensor by using calligraphic capital letter, such as $\AM$. Suppose $\AM$ is a tensor of size $I_1 \times I_2 \times \cdots \times I_L$, the order of $\AM$ is $L$ and the $l$th dimension (or mode) of $\AM$ is of size $I_l$. In addition, we denote the index of a single entry within a tensor by subscripts, such as $\AM_{i_1, \ldots, i_L}$.

\begin{definition}
The \textbf{scalar product} $\langle \AM, \BM \rangle$ of two tensors $\AM, \BM \in \RB^{I_1 \times I_2 \times \cdots \times I_L}$ is defined as $\langle \AM, \BM \rangle = \sum_{i_1} \cdots \sum_{i_L}  \AM_{i_1, \ldots, i_L} \BM^*_{i_1, \ldots, i_L}$, where $*$ denotes complex conjugation. Furthermore, the \textbf{Frobenius norm} of a tensor $\AM$ is defined as $\|A\|_{F} = \sqrt{\langle \AM, \AM \rangle}$.
\end{definition}

\begin{definition}
The \textbf{$l$-mode product} of a tensor $\AM \in \RB^{I_1 \times I_2 \times \cdots \times I_L}$ and a matrix $\U \in \RB^{J_l \times I_l}$ is an
$I_1\times \cdots \times I_{l-1} \times J_l \times I_{l+1} \times \cdots \times I_L$
tensor denoted as $\AM \times_l \U$, where the corresponding entries are given by $(\AM \times_l \U)_{i_1, \ldots, i_{l-1}, j_l, i_{l+1}, \ldots, i_L} = \sum_{i_l} \AM_{i_1, \ldots, i_{l-1}, i_l, i_{l+1}, \ldots, i_L} \U_{j_l i_l}$.
\end{definition}

\begin{definition}
Let $\AM$ be an $I_1 \times I_2 \times \cdots \times I_L$ tensor and ($\pi_1, \ldots, \pi_{L-1}$) be any permutation of the entries of the set $\{1, \ldots, l-1, l+1, \ldots, L\}$. The \textbf{$l$-mode unfolding} of the tensor $\AM$ into an $I_l \times \prod^{L-1}_{k = 1} I_{\pi_k}$ matrix, denoted as $\A^{(l)}$, is defined by $\AM \in \RB ^{I_1 \times I_2 \times \cdots \times I_L} \Rightarrow_l \A^{(l)} \in \RB^{I_l \times \prod^{L-1}_{k = 1} I{\pi_k}}$, where $ \A^{(l)}_{i_l j} = \AM_{i_1, \ldots, i_L}$ with
$j = 1+ \sum^{L-1}_{k = 1} (i_{\pi_k} - 1)  \prod^{k-1}_{\hat{k} = 1} I{\pi_{\hat{k}}}$.
\end{definition}

\begin{definition}
The \textbf{multi-linear rank} of a tensor is a set of nonnegative numbers, $(r_1, r_2, \ldots, r_L)$, such that
\begin{eqnarray}
r_l = \hbox{dim}(R(\A^{(l)}) = \hbox{rank}(\A^{(l)}), \  l = 1, 2, \ldots, L  \nonumber
\end{eqnarray}
where $R(\A) = \{ \f | \f = \A \z \}$ is the range space of the matrix $\A$, and rank($\A$) is the matrix rank.
\end{definition}

Multi-linear rank of tensors is elegantly discussed in~\citep{Illposed}, as well as other rank concepts. In this paper, we only focus on multi-linear rank of tensors and call it ``rank" for short.

\subsection{Formulation of LMLRTA}

As pointed out by researchers in the area of cognitive psychology that humans learn based on the similarity of examples~\citep{Rosch}, here, we formulate our model based on the local similarity between tensor data. In addition, thanks to the existence of many ``teachers", we can generally obtain the categorical information of the examples before or during learning. Take, for example, the moment when someone introduces an individual to his friend. His friend will probably remember the name of the individual first, and then her or his face and voice. In this case, name of the individual corresponds to a categorical label, whilst her or his face and voice are features to perceive. In the same way, we formulate our learning model in a supervised scheme.

Given a set of $N$ tensor data, $\{\AM_1, \ldots, \AM_N\} \in \RB^{I_1 \times \ldots \times I_L}$, with the associated class labels $\{\y_1, \ldots, \y_N\} \in \{1, 2, \ldots, C\}$, where $L$ is the order of the tensors and $C$ is the number of classes, we learn $L$ low rank projection matrix $\U_l \in \RB^{J_l \times I_l}\ (J_l \leq I_l, l = 1, \ldots, L)$,  such that $N$ embedded data points $\{\BM_1, \ldots, \BM_N\} \in \RB^{J_1 \times \ldots \times J_L}$ can be obtained as $\BM_i = \AM_i \times_1 \U_1 \times_2 \ldots \times_L \U_L$. The objective function can be written as
\begin{eqnarray}
\min && \LM(\lambda, \mu, \U_{l}|_{l = 1}^L) = \mu \sum_{l = 1}^L \hbox{rank}(\U_l) + \frac{\lambda}{2NL} \sum\limits_{i, j} \eta_{ij} \|\BM_i - \BM_j \|_F^2 \nonumber \\
&+& \frac{1}{2NL} \sum\limits_{i, j, p} \eta_{ij}(1 - \psi_{ij}) [1+\|\BM_i - \BM_j \|_F^2 - \|\BM_i - \BM_p \|_F^2]_\dag
\end{eqnarray}

where $\hbox{rank}(\U_l)$ is the rank of matrix $\U_l$, $\|\AM\|_F$ is the Frobenius norm of a tensor $\AM$, and $[z]_\dag = \max(0, z)$ is the so-called hinge loss, which is aimed at maximizing the margin between classes. If $\AM_i$ and $\AM_j$ have the same class label, and $\AM_i$ is one of the $k_1$-nearest neighbors of $\AM_j$ or $\AM_j$ is one of the $k_1$-nearest neighbors of $\AM_i$, then $\eta_{ij} = 1$, otherwise $\eta_{ij} = 0$. If $\AM_i$ and $\AM_j$ have different class labels, and $\AM_i$ is one of the $k_2$-nearest neighbors of $\AM_j$ or $\AM_j$ is one of the $k_2$-nearest neighbors of $\AM_i$, then $\psi_{ij} = 0$, otherwise $\psi_{ij} = 1$, i.e.
\begin{eqnarray}
\psi_{ij} =
\left\{
  \begin{array}{ll}
    0, & \hbox{$\y_i \neq \y_j$ and $\AM_j \in \mathfrak{N}_{k_2} (\AM_i)$ or $\AM_i \in \mathfrak{N}_{k_2} (\AM_j)$;} \\
    1, & \hbox{otherwise,}
  \end{array}
\right.
\end{eqnarray}
where $\mathfrak{N}_k (\AM_i)$ stands for $k$-nearest neighbor of $\AM_i$. Like the binary matrix $\{\eta_{ij}\}$ , the matrix $\{\psi_{ij}\}$ is fixed and does not change during learning.

The minimization of the first term of the objective function, $\sum_{l = 1}^L \hbox{rank}(\U_l)$, is to learn low rank $\U_l$'s and further the low dimensional representations of the tensors. The second term of the objective function is to enforce the neighboring data in each class to be close in the low dimensional tensor subspace. It can be considered as a graph Laplacian-parameterized loss function with respect to the low dimensional embeddings~\citep{Chung1997, BelkinN03, Tenenbaum11032011}, where each node corresponds to one tensor datum in the given data set. For each tensor datum $\AM_i$, the hinge loss in the third term will be incurred by a differently labeled datum within $k_2$-nearest neighbors of $\AM_i$, if whose distance to $\AM_i$ does not exceed, by 1, the distance from $\AM_i$ to any of its $k_1$-nearest neighbors within the class of $\AM_i$. This third term thereby favors projection matrices in which different classes maintain a large margin of distance. Furthermore, it encourages nearby data in different classes far apart in the low dimensional tensor subspace.

$\hbox{rank}(\U_l)$ is a non-convex function with respect to $\U_l$ and difficult to optimize. Following recent work in matrix completion~\citep{CandesT10, Candes2012}, we replace it with its convex envelope --- the nuclear norm of $\U_l$, which is defined as the sum of its singular values, i.e. $\|\U_l\|_* = \sum\limits_{s = 1}^r \sigma_s(\U_l)$ , where $\sigma_s(\U_l)$'s are the singular values of $\U_l$, and $r$ is the rank of $\U_l$. Thus, the resulting formulation of our model can be written as
\begin{eqnarray}\label{eq:obj1}
\min && \LM(\lambda, \mu, \U_{l}|_{l = 1}^L) = \mu \sum_{l = 1}^L \|\U_l\|_* + \frac{\lambda}{2NL} \sum\limits_{i, j} \eta_{ij} \|\BM_i - \BM_j \|_F^2 \nonumber \\
&+& \frac{1}{2NL} \sum\limits_{i, j, p} \eta_{ij}(1 - \psi_{ip}) [1+\|\BM_i - \BM_j \|_F^2 - \|\BM_i - \BM_p \|_F^2]_\dag
\end{eqnarray}

Since Problem (\ref{eq:obj1}) is not convex with respect to $\U_l$, we transform it into a convex problem with respect to $\W_l = \U_l^T \U_l$. Meanwhile, using the slack variables, Problem (\ref{eq:obj1}) can be rewritten as
\begin{footnotesize}
\begin{eqnarray}\label{eq:obj2}
\min && \LM(\lambda, \mu, \xi, \W_l|_{l = 1}^L) = \mu \sum_{l = 1}^L \|\W_l\|_* + \frac{\lambda}{2NL} \sum\limits_{i, j} \eta_{ij} \tr((\Y^{(l)}_i - \Y^{(l)}_j) (\Y^{(l)}_i - \Y^{(l)}_j)^T \W_l) \nonumber \\
&+& \frac{1}{2NL} \sum\limits_{i, j, p} \eta_{ij}(1 - \psi_{ip}) \xi_{ijp}  \nonumber \\
s.t. && \tr((\Y^{(l)}_i - \Y^{(l)}_p) (\Y^{(l)}_i - \Y^{(l)}_p)^T \W_l) - \tr((\Y^{(l)}_i - \Y^{(l)}_j) (\Y^{(l)}_i - \Y^{(l)}_j)^T \W_l) \geq 1 - \xi_{ijp}, \nonumber \\
&& \xi_{ijp} \geq 0, \quad i, j, p = 1, 2, \ldots, N,
\end{eqnarray}
\end{footnotesize}
where $\Y^{(l)}_i$ is the $l$-mode unfolding matrix of the tensor $\YM_i = \A_i \times _1 \U_1 \times _2 \ldots \times_{l-1} \U_{l-1} \times_{l+1} \U_{l+1} \times _{l+2} \ldots \times_L \U_L$. For the second term of the objective function and the first constraint in Problem (\ref{eq:obj2}), we have used the property of the trace function: $\tr(\U_l(\Y^{(l)}_i - \Y^{(l)}_j) (\Y^{(l)}_i - \Y^{(l)}_j)^T\U_l^T) = \tr((\Y^{(l)}_i - \Y^{(l)}_j) (\Y^{(l)}_i - \Y^{(l)}_j)^T\U_l^T\U_l)$.

The equivalence between Problem (\ref{eq:obj1}) and Problem (\ref{eq:obj2}) can be guaranteed by the following lemma.
\begin{lemma}
Based on the notation above, Problem (\ref{eq:obj1}) and Problem (\ref{eq:obj2}) are equivalent.
\end{lemma}
\proof
Based on simple computation, we know that the second term of the objective function in Problem (\ref{eq:obj1}) is equal to that in Problem (\ref{eq:obj2}), while the third term of the objective function in Problem (\ref{eq:obj1}) is equivalent to that in Problem (\ref{eq:obj2}) with the constraints. As $\sigma_s(\U_l) = \sqrt{\sigma_s(\W_l)}$, the optimal solution of Problem (\ref{eq:obj1}) must correspond to the optimal solution of Problem (\ref{eq:obj2}), and vice versa, where $\sigma_s(\U_l)$ and $\sigma_s(\W_l)$ are the singular values of $\U_l$ and $\W_l$, respectively. Thus, the lemma is proved.
\endproof

Problem (\ref{eq:obj2}) is not jointly convex with respect to all the $\W_l$'s. However, it's convex with respect to each of them. This is guaranteed by the following lemma.
\begin{lemma}\label{th:convex}
Problem (\ref{eq:obj2}) is convex with respect to each $\W_l$.
\end{lemma}
\proof
First, the nuclear norm of $\W_l$, $\|\W_l\|_*$, is a convex function with respect to $\W_l$. Second, the other terms of the objective function and the constraints in Problem (\ref{eq:obj2}) are all linear function with respect to $\W_l$. Hence, Problem (\ref{eq:obj2}) is convex with respect to each $\W_l$.
\endproof

\begin{remark}[Relation to previous works]
\begin{itemize}
  \item[1)] LMLRTA can be considered as a supervised multi-linear extension of locality preserving projections (LPP)~\citep{HeN03}, in that the second term of the objective function in Problem (\ref{eq:obj2}) forces neighboring data in a same class to be close in the low dimensional tensor subspace;
  \item[2)] LMLRTA can also be considered as a reformulation of tensor marginal Fisher analysis (TMFA)~\citep{YanXZZYL07}. However, TMFA is not guaranteed to converge to a local optimum of the optimization problem~\citep{WangYHT07}, but LMLRTA is guaranteed as proved in Section~\ref{sec:optimization};
  \item[3)] For Problem (\ref{eq:obj2}), we can consider it as a variant of the Large Margin Nearest Neighbor (LMNN) algorithm~\citep{WeinbergerBS05} for distance metric learning in tensor space. Moreover, we can learn \textbf{low rank} distance matrices via the formulation of Problem (\ref{eq:obj2}), which the LMNN algorithm is not endowed;
  \item[4)] In contrast to previous approaches for tensor dimensionality reduction, which can only learn project matrices with pre-specified dimensionality of the low dimensional representations, LMLRTA can automatically learn the dimensionality of the low dimensional representations from the given data. This will be shown in Section~\ref{sec:optimization}.
  \item[5)] Unlike deep neural network models~\citep{HintonOT06, BengioLPL06, VincentLLBM10}, which simulate human brain's hierarchical structure, LMLRTA mimics the way of human perception. On one hand, LMLRTA can take any order of tensors as input, but most deep neural networks only take vectorized representations of data. On the other hand, with large number of parameters, the learning of deep neural network models in general needs many training data. If the size of the training set is small, deep neural network models may fail to learn the intrinsic structure of data. However, in this case, LMLRTA can perform much better than deep neural network models. Experimental results in Section~\ref{sec:exres} demonstrate this effect.
\end{itemize}
\end{remark}

\subsection{Optimization}\label{sec:optimization}

Similar to previous approaches on tensor dimensionality reduction~\citep{DYYeung, WangYHT07}, here we solve Problem (\ref{eq:obj2}) using an iterative optimization algorithm. In each iteration, we refine one projection matrix by fixing the others. Here, for each $\W_l$, problem (\ref{eq:obj2}) is a semi-definite programming problem, which can be solved using off-the-shelf algorithms, such as SeDuMi\footnote{http://sedumi.ie.lehigh.edu/} and CVX~\citep{gb08}. However, the computational cost of semi-definite programming approaches is in general very high. Here, we solve the problem by means of a modified fixed point continuation (MFPC) method~\citep{MaGC11}.

MFPC is an iterative optimization method. In the $t$-th iteration, it involves two alternating steps:
\begin{itemize}
  \item[a)] Gradient step:  $\Z_l^t = \W_l^t - \tau g(\W_l^t)$;
  \item[b)] Shrinkage step: $\W_l^{t+1} = S_{\tau \mu}(\Z_l^t)$.
\end{itemize}

In the gradient step, $g(\W_l^t)$ is the sub-gradient of the objective function in problem (\ref{eq:obj2}) with respect to $\W_l^t$ (excluding the nuclear norm term), and $\tau$ is the step size. Here, we can express $\xi_{ijp}$ as a function with respect to $\W_l^t$:
\begin{footnotesize}
\begin{eqnarray}
&& \xi_{ijp}(\W_l^t) = [1 + \tr((\Y^{(l)}_i - \Y^{(l)}_j) (\Y^{(l)}_i - \Y^{(l)}_j)^T \W_l) - \tr((\Y^{(l)}_i - \Y^{(l)}_p) (\Y^{(l)}_i - \Y^{(l)}_p)^T \W_l)]_\dag \nonumber \\
&& i, j, p = 1, 2, \ldots, N.
\end{eqnarray}
\end{footnotesize}
Note that the hinge loss is not differentiable, but we can compute its sub-gradient and use a standard descent algorithm to optimize the problem. Thus we can calculate $g(\W_l^t)$ as
\begin{footnotesize}
\begin{eqnarray}
&& g(\W_l^t) = \frac{\lambda}{2NL} \sum\limits_{i, j} \eta_{ij} (\Y^{(l)}_i - \Y^{(l)}_j) (\Y^{(l)}_i - \Y^{(l)}_j)^T \nonumber \\
&& + \frac{1}{2NL} \sum\limits_{\{i, j, p\} \in \mathfrak{S} } \eta_{ij}(1 - \psi_{ip}) ((\Y^{(l)}_i - \Y^{(l)}_j) (\Y^{(l)}_i - \Y^{(l)}_j)^T - (\Y^{(l)}_i - \Y^{(l)}_p) (\Y^{(l)}_i - \Y^{(l)}_p)^T),
\end{eqnarray}
\end{footnotesize}
where $\mathfrak{S}$ is the set of triplets whose corresponding slack variable exceeds zero, i.e., $\xi_{ijp}(\W_l^t) > 0$.

In the shrinkage step, $S_{\tau \mu}(\Z_l^t) = \V \max\{\0, \mathbf{\Lambda} - \hbox{diag}(\tau \mu)\} \V^T$ is a matrix shrinkage operator on $\Z_l^t = \V \mathbf{\Lambda} \V^T$, where max is element-wise and  $\hbox{diag}(\tau \mu)$ is a diagonal matrix with all the diagonal elements set to $\tau \mu$. Here, since $\W_l^t$ is supposed to be a symmetric and positive semi-definite matrix, its eigenvalues should be nonnegative. Therefore, we adapt the eigenvalue decomposition method to shrink the rank of $\Z_l^t$. To this end, the shrinkage operator shifts the eigenvalues down, and truncates any eigenvalue less than $\tau \mu$ to zero. \emph{This step reduces the nuclear norm of $\W_l^t$. If some eigenvalues are truncated to zeros, this step reduces the rank of $\W_l^t$ as well}. In our experiments, we use relative error as the stopping criterion of the MFPC algorithm.

For clarity, we present the procedure of the MPFC algorithm in Algorithm~\ref{algo:MPFC}.
\begin{algorithm}[t]
\caption{The MPFC algorithm.}
\label{algo:MPFC}
\begin{algorithmic}[1]
\STATE \textbf{Input:}
\STATE \ \ \ \ $\lambda$, $T_{max}$, $\W_l^0$, $\bar{\mu} > 0$; \quad $\%$ $T_{max}$ is the maximum number of iterations.
\STATE \textbf{Initialization:}
\STATE \ \ \ \  $\mu_1 > \mu_2 > \ldots > \mu_K = \bar{\mu}$;
\STATE \textbf{Steps:}
    \FOR {$\mu = \mu_1, \mu_2, \ldots, \mu_K$}
        \WHILE {$t < T_{max}$ and not converge}
        \STATE Compute $\Z_l^t = \W_l^t - \tau g(\W_l^t)$ and eigenvalue decomposition of $\Z_l^t$, $\Z_l^t = \V \mathbf{\Lambda} \V^T$;
        \STATE Compute $\W_l^{t+1} = S_{\tau \mu}(\Z_l^t)$;
        \ENDWHILE
    \ENDFOR
\STATE \textbf{Output:}
\STATE \ \ \ \ The learned $\W_l$.
\end{algorithmic}
\end{algorithm}

For the convergence of the MFPC algorithm, we present a theorem as below.
\begin{theorem}\label{th:convergeMFPC}
For fixed $\W_k$, $k = 1, \ldots, l - 1, l + 1, \ldots, L$, the sequence $\{\W_l^t\}$ generated by the MPFC algorithm with $\tau \in (0, 2/\lambda_{max}(g(\W_l)))$ converges to the optimal solution, $\W_l^*$, of Problem (\ref{eq:obj2}), where $\lambda_{max}(g(\W_l))$ is the maximum eigenvalue of $g(\W_l)$.
\end{theorem}
The proof of this theorem is similar to that of theorem 4 in \citep{MaGC11}. A minor difference is, we use eigenvalue decomposition here instead of singular value decomposition as used in the proof of theorem 4 in \citep{MaGC11}. However, the derivation and results are the same.

Based on the above lemmas and Theorem~\ref{th:convergeMFPC}, we can have the following theorem on the convergence of our proposed method, LMLRTA.
\begin{theorem}\label{th:LMLRTA}
LMLRTA converges to a local optimal solution of Problem (\ref{eq:obj2}).
\end{theorem}
\proof
To prove Theorem~\ref{th:LMLRTA}, we only need to prove that the objective function has a lower bound, as well as the iterative optimization procedures monotonically decrease the value of the objective function.

First of all, it's easy to see that the value of the objective function in Problem (\ref{eq:obj2}) is always larger than or equal to 0. Hence, 0 is a lower bound of this objective function. Secondly, for the optimization of each $\W_l$, $l = 1, \ldots, L$, from Theorem~\ref{th:convergeMFPC}, we know that the MPFC algorithm minimizes the value of the objective function in Problem (\ref{eq:obj2}). Therefore, the iterative procedures of LMLRTA monotonically decrease the value of the objective function, and LMLRTA is guaranteed to converge to a local optimal solution of Problem (\ref{eq:obj2}).
\endproof

Based on Lemma~\ref{th:convex} and Theorem~\ref{th:LMLRTA}, we can easily obtain a corollary as below:
\begin{cor}
If the given data are one-order tensors, the LMLRTA algorithm converges to the optimal solution of Problem (\ref{eq:obj2}).
\end{cor}

\subsection{Generalization to new tensor data}

For the recognition of unseen test tensors, we employ the tensor Frobenius norm-based  $k$-nearest neighbor classifier as recognizer, in that it measures the local similarity between training data and test data in the low dimensional tensor subspace~\citep{Rosch}.

\section{Experiments}\label{sec:exres}

In this section, we report the experimental results obtained on two real world applications: object recognition and face recognition. Particularly, for the \emph{face recognition} task on the ORL data set, we used \emph{3D Gabor transformation} of the face images as input signals. This is mainly based on the fact that \emph{the kernels of the Gabor filters resemble the receptive field profiles of the mammalian cortical simple cells}~\citep{Daugman1}, which enhances our learning model to better mimic the way of human perception. In the following, we report the parameter settings and experimental results in detail.

\subsection{Parameter settings}\label{sec:settings}

To demonstrate the effectiveness of our method for the intrinsic representation learning and recognition, we conducted experiments on the COIL-20 data set\footnote{http://www.cs.columbia.edu/CAVE/software/softlib/coil-20.php.} and the ORL face data set\footnote{http://www.cl.cam.ac.uk/research/dtg/attarchive/facedatabase.html.}. The COIL-20 data set includes 20 classes of objects, and 72 samples within each class. The size of the images is $32 \times 32$. The ORL data set contains 400 images of 40 subjects, where each image was normalized to a size of $32 \times 32$. For each face image, we used 28 Gabor filters to extract textural features. To the end, each face image was represented as a $32 \times 32 \times 28$ tensor. On the COIL-20 data set, we used 5-fold cross validation to evaluate the performance of the compared methods. The average classification results were reported. As each subject only has 10 images in the ORL data set, we evaluated the compared methods based on the average over 5 times random partition of the data. Here, variety of scenarios --- different numbers of training data from each class, were tested.

To show the advantage of our proposed method, LMLRTA, we compared it with two classic vector representation-based dimensionality reduction approaches -- linear discriminant analysis (LDA)~\citep{fisher36lda} and marginal Fisher analysis (MFA)~\citep{YanXZZYL07}, one deep neural networks model called stacked denoising autoencoder (SDAE)~\citep{VincentLLBM10}, and two state-of-the-art tensor dimensionality reduction methods -- convergent multi-linear discriminant analysis (CMDA) and convergent tensor margin Fisher analysis (CTMFA)~\citep{WangYHT07}. For comparison, we also provided the classification results obtained in the original data space. In the LMLRTA algorithm, $k_1$ and $k_2$ were set to 7 and 15 respectively for the COIL-20 data set, while for the ORL data set, they were set to \emph{$\hbox{ntrain} - 1$}  and \emph{$2 \times \hbox{ntrain}$} respectively, where \emph{$\hbox{ntrain}$} is the number of training data from each class. Furthermore, $\lambda$ was selected from $\{0.001, 0.01, 0.1, 1, 10\}$, and the one resulting best classification result was used. For CMDA and CTMFA, we adopted the best setting learned by LMLRTA to specify the dimensionality of the low dimensional tensor subspace. We used the code of SDAE from a public deep learning toolbox\footnote{https://github.com/rasmusbergpalm/DeepLearnToolbox.}. For all the methods but SDAE, tensor Frobenius norm-based 1-nearest neighbor classifier was used for the recognition of test data.

\begin{figure*}[ht]
\begin{center}
\subfigure[The COIL-20 data set.]{\includegraphics[width=7.2cm, height=5cm]{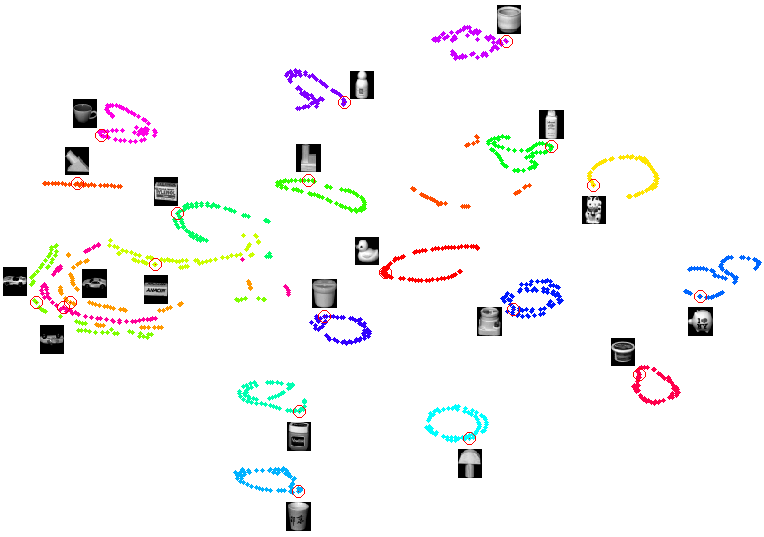}}  \hspace{0.5cm}
\subfigure[The ORL data set.]{\includegraphics[width=6.6cm, height=5cm]{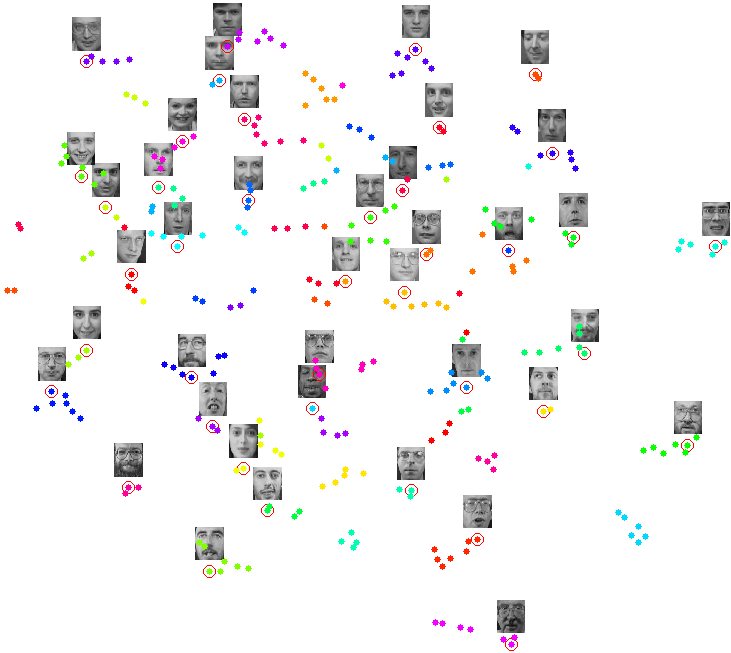}}
\end{center} \vskip -0.5cm
\caption{2D embeddings of the tensors from the COIL-20 data set and the ORL data set, where different classes are denoted with different colors. (a) Images from the COIL-20 data set. (b) Gabor transformation of the face images from the ORL data set. We can see that, in the \emph{original space} of these two data sets, some data of the same class are far apart, and at the same time, some are close to data of other classes.}
\label{fig_sim}
\end{figure*}

\subsection{Visualization}

Figure~\ref{fig_sim} (a) and Figure~\ref{fig_sim} (b) illustrate the 2D embeddings of the object images from the COIL-20 data set and that of the 3D Gabor transformation of the face images from the ORL data set, respectively. The t-distribution-based stochastic neighbor embedding (t-SNE) algorithm~\citep{Maaten} was employed to learn these 2D embeddings, where the distances between data were measured based on tensor Frobenius norm. From Figure~\ref{fig_sim} (a) and Figure~\ref{fig_sim} (b), we can see that, in the original space of these two data sets, most of the classes align on a sub-manifold embedded in the ambient space. However, for some classes, the data are scattered in a large area of the data space, and alternatively, close to data of other classes. As a result, similarity-based classifiers may predict the labels of some unseen data incorrectly in both of these two original representation spaces. Hence, it's necessary to learn the intrinsic and informative representations of the given tensor data.

\begin{figure*}[ht]
\begin{center}
\subfigure[The COIL-20 data set.]{\includegraphics[width=7.2cm, height=5cm]{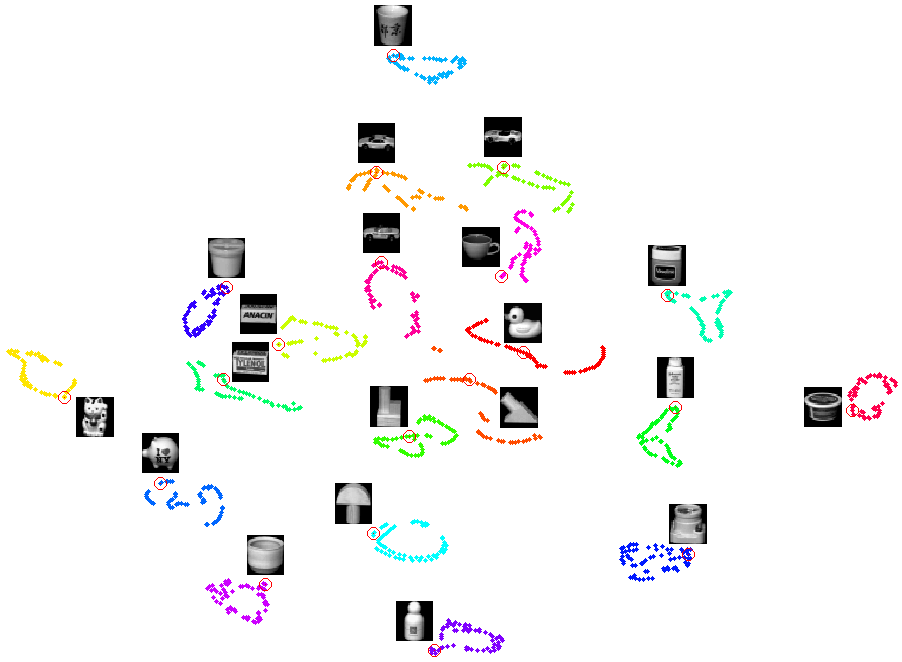}}  \hspace{0.5cm}
\subfigure[The ORL data set.]{\includegraphics[width=6.6cm, height=5cm]{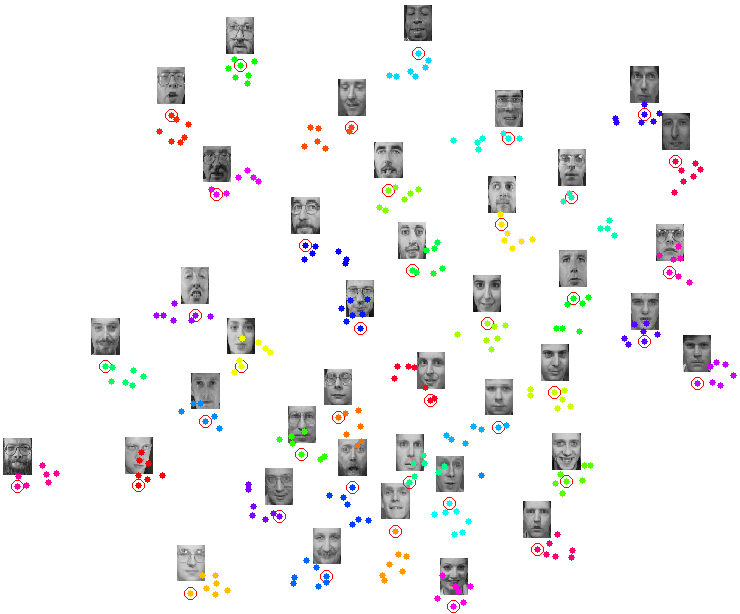}}
\end{center} \vskip -0.5cm
\caption{2D embeddings of the low dimensional tensor representations for the COIL-20 and the ORL data set. LMLRTA was used to learn the low dimensional tensor representations. (a) Corresponding low dimensional tensor representations of the images shown in Figure~\ref{fig_sim} (a). (D) Corresponding low dimensional tensor representations of the 3D Gabor transformation of the face images shown in Figure~\ref{fig_sim} (b). We can see that, in the learned low dimensional tensor subspace by LMLRTA, the data points of the same class are close to each other, while data of different classes are relatively far apart.}
\label{fig_sim2}
\end{figure*}

Figure~\ref{fig_sim2} (a) and Figure~\ref{fig_sim2} (b) illustrate the 2D embeddings of the low dimensional tensor representations for the COIL-20 and the ORL data set, respectively. Here, LMLRTA was used to learn the low dimensional tensor representations, while the t-SNE algorithm was used to generate the 2D embeddings. It is easy to see, LMLRTA successfully discovered the manifold structure of these two data sets. In both Figure~\ref{fig_sim2} (a) and Figure~\ref{fig_sim2} (b), the similarity between data of the same class are faithfully preserved, whilst the discrimination between classes are maximized.

\begin{figure*}[ht]
\begin{center}
\subfigure[CMDA]{{\includegraphics[width=1.75in, height=1.75in]{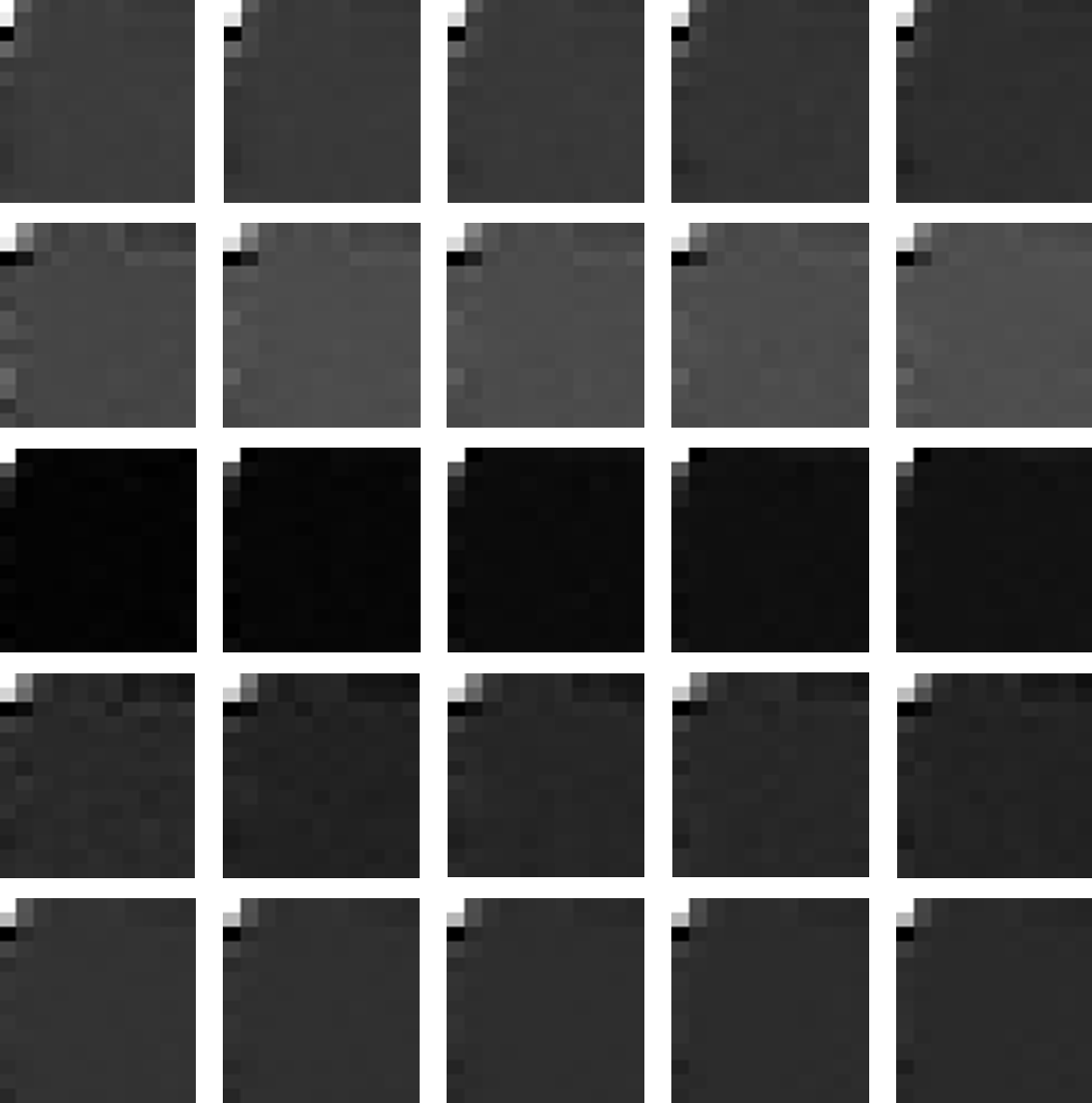}}}
\subfigure[CMFA]{\includegraphics[width=1.75in, height=1.75in]{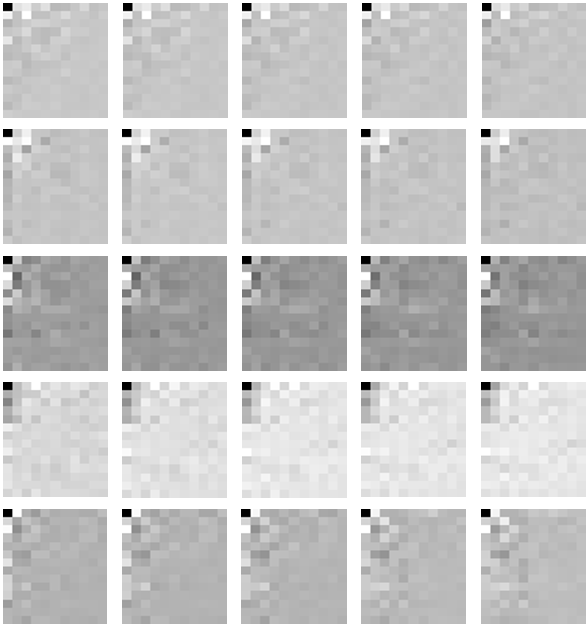}}
\subfigure[LMLRTA]{\includegraphics[width=1.75in, height=1.75in]{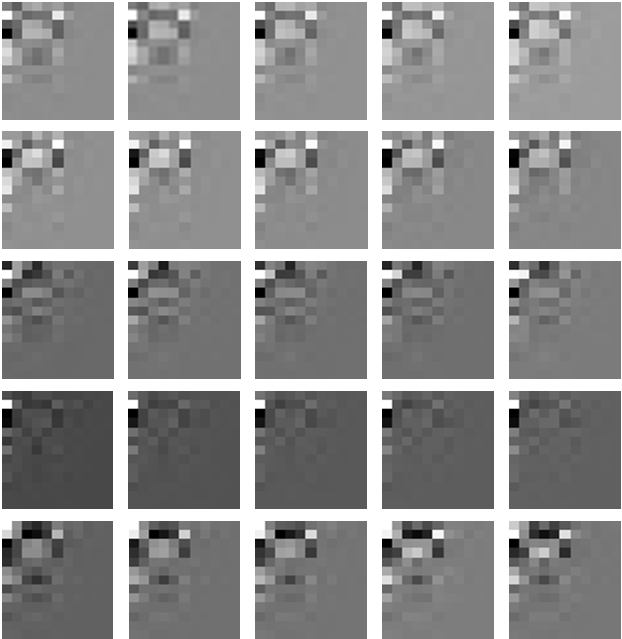}}
\end{center} \vskip -0.5cm
\caption{Learned low dimensional tensor representations of images from the COIL-20 data set. (a) Those learned by CMDA. Here, each row shows the low dimensional representations of images from one class. Five classes are totally shown. (b) Low dimensional tensor representations of same images as in (a) learned by CTMFA. (c) Low dimensional tensor representations of same images as in (a) learned by LMLRTA. We can see that, in the learned low dimensional tensor subspace, all three methods preserve the similarity between data within each class faithfully. However, classification results show that, the discrimination between classes in the tensor subspace learned by LMLRTA is better than those learned by CMDA and CTMFA.}
\label{fig_sim3}
\end{figure*}

Figure~\ref{fig_sim3} shows some low dimensional tensor representations of the images from the COIL-20 data set, which were learned by CMDA (a), CTMFA (b) and LMLRTA (c), respectively. Five classes were randomly selected, and low dimensional representations of five images were further randomly selected to show for each class. Particularly, in each sub-figure of Figure~\ref{fig_sim3}, each row shows the low dimensional tensor representations of images from one class. In contrast to the dimensionality of the original image, $32 \times 32$, the dimensionality of the low dimensional representations here is $12 \times 11$. We can see that, all three methods can preserve the similarity between data of the same class faithfully. However, the discrimination between classes in the low dimensional tensor subspace learned by LMLRTA is much better than those learned by CMDA and CTMFA. Recognition results shown in Section~\ref{sec:COILrec} also demonstrate this observation.

\subsection{Object recognition results on the COIL-20 data set (2D tensors)}\label{sec:COILrec}

In this experiment, we compare LMLRTA with some related approaches on the object recognition application. The compared approaches include LDA, MFA, SDAE, CMDA, CTMFA and classification in the original space. We implemented experiment on the COIL-20 data set. To conduct this experiment, we empirically tested the dimensionality of the LDA subspace and that of the MFA subspace, and fixed them to 19 and 33, respectively. For the SDAE algorithm, we used a 6-layer neural network model. The sizes of the layers were 1024, 512, 256, 64, 32 and 20, respectively. For LMLRTA, CMDA and CTMFA, we just followed the settings as introduced in Section~\ref{sec:settings}.

\begin{figure*}[ht]
\begin{center}
\includegraphics[width=4.5in, height=3in]{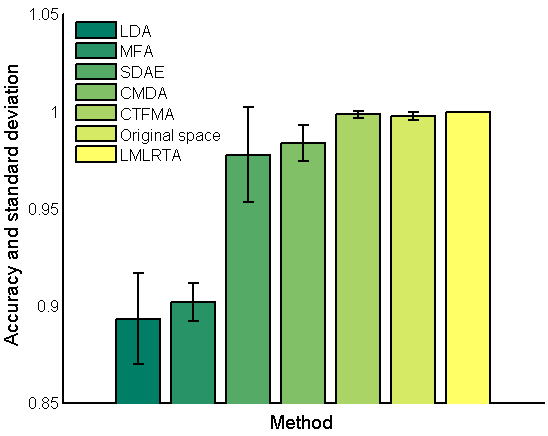}
\end{center}
\caption{Classification results obtained by the compared methods on the COIL-20 data set. Note that LMLRTA obtained 100\% accuracy over all five folds cross validation, but SDAE only provided a 97.8\% accuracy for that. }
\label{fig_sim4}
\end{figure*}

Figure~\ref{fig_sim4} shows the classification accuracy and standard deviation obtained by the compared methods. It is easy to see that, LMLRTA performed best among all the compared methods, as it achieved 100\% accuracy over all 5 folds of cross validation. Due to the loss of local structural information of the images, vector representation-based approaches, LDA and MFA, performed worst on this problem. Because of the limitation of training sample size, deep neural network model, SDAE, can not outperform LMLRTA on this problem and shew a large standard deviation. State-of-the-art tensor dimensionality reduction approaches, CMDA and CTMFA, can converge to a local optimal solution of the learning problem, but not perform as well as LMLRTA.

\begin{figure*}[ht]
\begin{center}
\includegraphics[width=4.5in, height=3in]{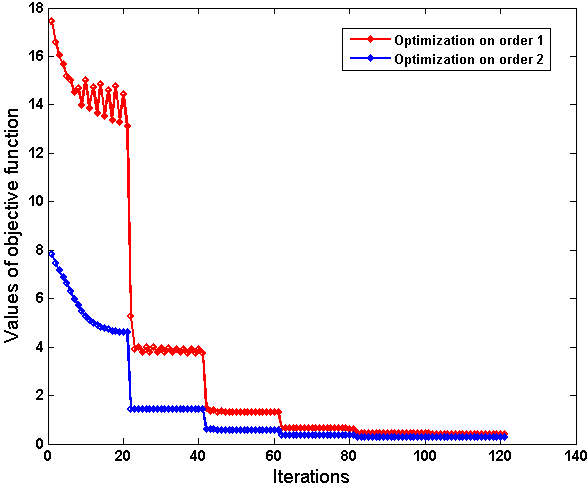}
\end{center}
\caption{The optimization for the two orders of one projection matrix. These two curves show that the MPFC algorithm can converge to a stationary point of the optimization problem.}
\label{fig_sim5}
\end{figure*}

To show the convergence process of the MPFC algorithm during learning of the projection matrices, Figure~\ref{fig_sim5} illustrates the values of the objective function against iterations during the optimization of LMLRTA on the COIL-20 data set. As we can see, the MPFC algorithm converges to a stationary point of the problem as the iteration continues.

\subsection{Face recognition results on the ORL data set (3D tensors)}

\begin{figure*}[ht]
\begin{center}
\includegraphics[width=4.5in, height=3in]{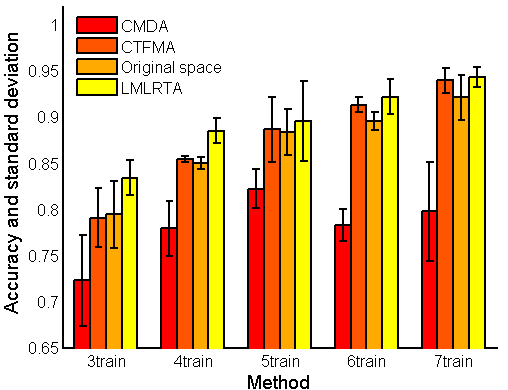}
\end{center}
\caption{Recognition results for the ORL face images.}
\label{fig_sim6}
\end{figure*}

Figure~\ref{fig_sim6} shows the classification accuracy and standard deviation obtained on the ORL data set. Due to high computational complexity problems of LDA, MFA and SDAE (the vector representations of the tensors is of dimensionality $32 \times 32 \times 28 = 28672$), here we only compared LMLRTA to CMDA, CTMFA and the classification in the original data space. From Figure~\ref{fig_sim6}, we can see that LMLRTA consistently outperforms the compared convergent tensor dimensionality reduction approaches. More importantly, as LMLRTA gradually reduces the ranks of the projection matrices during optimization, it can learn the dimensionality of the intrinsic low dimensional tensor space automatically from data. However, for traditional tensor dimensionality reduction algorithms, the parameter must be manually specified before they can be applied. This may result in unsatisfactory results on the applications.

\section{Conclusion}

In this paper, we propose a supervised tensor dimensionality reduction method, called large margin low rank tensor analysis (LMLRTA). LMLRTA can be utilized to automatically and jointly learn the dimensionality and representations of low dimensional embeddings of tensors. This property makes it an effective simulation of the way how human brain represents perceived signals. To recognize new coming data, we employ similarity based classifiers in the learned tensor subspace, which corresponds to the recognition procedure of human brain~\citep{Rosch}. Experiments on object recognition and face recognition show the superiority of LMLRTA over classic vector representation-based dimensionality reduction approaches, deep neural network models and existing tensor dimensionality reduction approaches. In future work, we attempt to extend LMLRTA to the scenarios of transfer learning~\citep{PanY10} and active learning~\citep{Cohn94}, to simulate the way how human brain transfers knowledge from some source domains to a target domain, and the way how human brain actively generates questions and learns knowledge. Furthermore, we plan to combine LMLRTA with deep neural networks~\citep{lecun01a} and non-negative matrix factorization models~\citep{Lee}, to solve challenging large scale problems.

\section*{Acknowledgments}

We thank the Social Sciences and Humanities Research Council of Canada (SSHRC) as well as the Natural Sciences and Engineering Research Council of Canada (NSERC) for their financial support.

\bibliographystyle{apa}
\bibliography{refs}

\end{document}